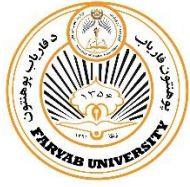

**Ministry of Higher Education**
**Faryab University**
**Literature and Humanities Faculty**
**English Department**

**Principle Methods of Rendering Non-equivalent Words from Uzbek and Dari to Russian and English**

**Mohammad Ibrahim Qani**
**Senior Teaching Assistant (Pohanmal)**

**Mobile: +93794555300**

**E-mail: ibrahim.qani@gmail.com & ibrahim.qani2021@faryab.edu.af**

**2025**

**Abstract**

Translating words which do not have equivalent in target language are not easy and finding proper equivalent of those words are very important to render correctly and understandably, the article defines some thoughts and ideas of scientists on the common problems of non-equivalent words from English to Russian language and includes English and Russian examples and ideas of certain scientist. The English language is worldwide spoken and there are 1.35 billion English speakers and over 258 million Russian speakers according to the 2021's statistics. Inevitably, these billions of speakers around the world have connection and they may have deal in different criteria. In order to understand one another they need to have a pure and fully-understood language. These pure languages understanding directly relates to translation knowledge where linguists and translators need to work and research to eradicate misunderstanding. Misunderstandings mostly appear in non-equivalent words because there are different local and internal words like food, garment, cultural and traditional words and others in every notion. Truly, most of these words do not have equivalent in the target language and these words need to be worked and find their equivalent in the target language to fully understand the both languages. The purpose of this research is to introduce the methods of rendering non-equivalent words professionally from the source language to the target language and this research has been completed using library-based research. However, some of these non-equivalent words are already professionally rendered to the target language but still there many other words to be rendered. As a result, this research paper includes different ways and rules of rendering non-equivalent words from source language to the target language and 25 non-eqvalent words have been rendered from Dar & Uzbek into English and Russian languages.

**Methods of Research**

In this article, I used the APA style and the usual and old research method which is called Library Research (data collection), online libraries, PDF books, available research papers in reliable journals and websites, and some other sources. This article covers the analysis of rendering some sort of technical and scientific words from the source language into the target language.

**Introduction**

Translation and interpreting play basic roles in today's world in order to communicate properly and understand one another. However, the development of technology has connected people around the world and people can easily contact one another using different devices, which are connected to the internet. As we know learning different languages is very important in today's world, and most people around the world try to learn different languages, because people will extremely need different languages in the near future in order to have clear communication.

Translation is a complex subject. Translating non-equivalent words is one of the significant subjects of translation, and there are different ways and methods to render such words of the source language in the target language. I will briefly describe some useful and effective methods with examples and by using them we can easily render words that do not have equivalence or non-equivalent words in different target languages.

**Transcription**

According to Chomsky, the possibility to translate is unlimited as far as "closed" texts are concerned - texts that can be interpreted in a single way, without connotations, i.e. a minimal part of real texts (CHOMSKY, 1976, p. 35).

Examples:

| Dari (Persian) | Uzbek | Russian | English |
|---|---|---|---|
| Pakora (پکوره) | Pakora (پکوره) | Пакора | Pakora |
| Ghalghla Shaitan (غلغله شیطان) | Ghalghla Shaitan (غلغله شیطان) | Гальгла Щайтан | Ghalghla Shaitan |

**Transliteration**

The act or process of writing words using a different alphabet:
- Their texts aimed to produce a phonetic transliteration of the dialect, using the main languages of the day.
- Transliteration helps people speak a language by showing the pronunciation in the language they understand (Combrage, 2023, p. 5).

Examples:

| Dari (Persian) | Uzbek | Russian | English |
|---|---|---|---|
| Halva (حلوا) | Halva (حلوه) | Хальва / Халва | Halva |
| Umaj (اوماج) | Umach (اوماچ) | Умадж | Umaj |

**Loan Translation**

In linguistics, a calque (/kælk/) or loan translation is a word or phrase borrowed from another language by literal word-for-word or root-for-root translation. When used as a verb, "to calque" means to borrow a word or phrase from another language while translating its components, so as to create a new lexeme in the target language (Shapiro, 2013, p. 53).

A *loan translation* is a compound in English (for example, *superman*) that literally translates a foreign expression (in this example, German *Übermensch*), word for word (Nordquist, 2018, p. 21).

Examples:

| Dari (Persian) | Uzbek |
|---|---|
| Tushla Bazi (تشله بازی) | Tushla Oyeni (تشله اوینی) |
| Toop Danda (توپ دنده) | Qachar Toop (قاچر توپ) |

**Approximate Translation**

Approximate Translation is a poetic and practical rumination on how to incorporate what makes a city a city (Crisman, 2024, p. 35).

Approximate translation is used when the source language has a word denoting a similar or close notion to the word used in the source language. If the target language doesn't have such a word then we cannot use approximate translation. We have to be very sensitive in using this method, before using this method we have to search similarity of words in the target language.

Examples:

| Dari (Persian) | Russian | English |
|---|---|---|
| Sheer Brenj (شیر برنج) | Рисовая Каша | Rice Pudding |
| Ash Qashoqi (آش قاشقی) | Афганский лагман | Nudel/Noodle Soup |

**Descriptive Translation**

The use of a description to translate a term or a phrase in the source by characterizing it instead of translating it directly" (Darwish, 2010, p. 142).

This type of rendering of non-equivalent words is also called **explanation**. By using this method we have to try to describe or explain the meaning of the source word or phrase in the target language (Zuckermann, 2003, p. 14).

National – cultural component of non-equivalent vocabulary being semantic heterogeneous micro component find its expression in semantic structure of a word by obligatory sem "locality", "ethnic accessory" and optional sem "historical relatedness", "social-political activities", "socio-cultural information", "confessional accessory" (Zokirova Sohiba, 2016, p. 71).

Examples:

| Dari (Persian) | Russian | English |
|---|---|---|
| Showlah-e-zard (شله زرد) | (Желтая Рисовая Каша) (Тыквенная Каша без | Yellow Rice Porridge |

**List of Non-equivalent words**

Non-equivalent Word Combinations of Educational Type. Implementation of our proposed lexicographical project can become highly relevant, as the need for its preparation is caused by the necessity of learning English as a mean of communication without isolation from the culture of the target language (Elena, 2015, p. 388).

Here, I have rendered 16 non-equivalent words from Dari and Uzbek languages to English and Russian languages using the above types of translation methods and the terms have been explained and clearly shown by their images.

| **Non-equivalences** |
|---|
| **Food** |

| Uzbek Language of Afghanistan | Uzbek language of Uzbekistan | Dari (Persian) | Russian | English |
|---|---|---|---|---|
| Ekipiazli (ایپکی پیازلی) | Ekipiazli Ovqat | Dupiaza (دوپیازه) | Дупияаза | Dupiaza |
| **Explanation:** This traditional food is also very popular and delicious; Dupiaza is a Dari (Persian) compound word which means {Du (Two) Piaz (Onion) two onions} Afghans cook this food mostly for weddings and parties. Ingredients: meat (veal or lamb), white onion, oil, black pepper, fresh pepper, vinegar, fat tail, and spices. ||||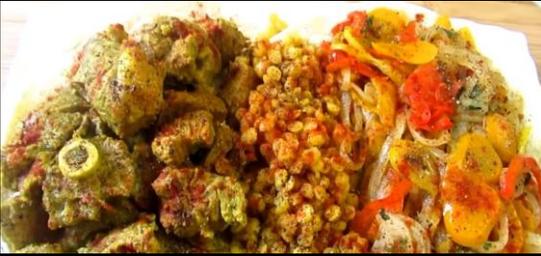|
| Sotli Brenj (سوتلی برنج) | Shirguruch | Sheer Brenj (شیر برنج) | Щирбриндж (Рисовая Каша) | Sheer Brenj (Rice Pudding) |
| **Explanation:** This food is mostly used at parties and most Afghan people prefer to eat it cold. This is a sweet and delicious dish. Ingredients: Milk, rice, sugar, water, cardamom, creamy milk and salt. ||||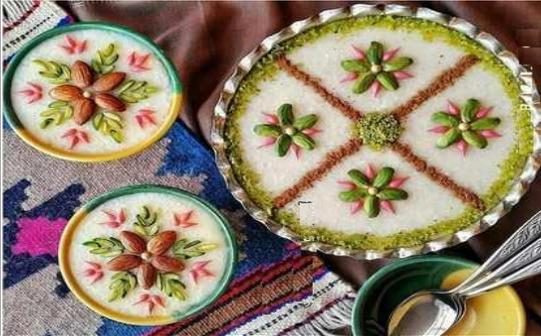|
| Umach (اوماچ) | O'moch | Umaj (اوماج) | Умадж | Umaj (soup) |
| **Explanation:** Umaj is also one of the ancient food which is mostly cooked and used in winter. It's cooked in the winter because it keeps the body so warm. Ingredients: Lentils, onion, salt, pea, bean, peas (coriander, mint, cabbage), beetroot, vinegar, oil, salt, pepper and turmeric. ||||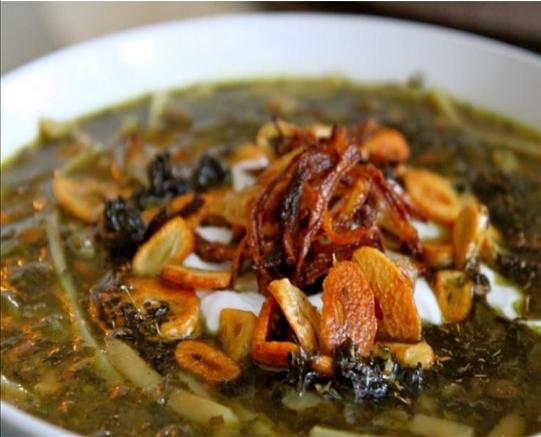|
| Ash Qouroot (آش قوروت) | Ugra Oshi | Ash Qoroot (آش قروت) | Ащ Курот | Ash Qoroot (Goat milk soup) |
| **Explanation:** This food has come to Afghanistan from Asian countries and some sources mentioned that it has come from Azerbaijan but still we don't have reliable sources to prove this claim. ||||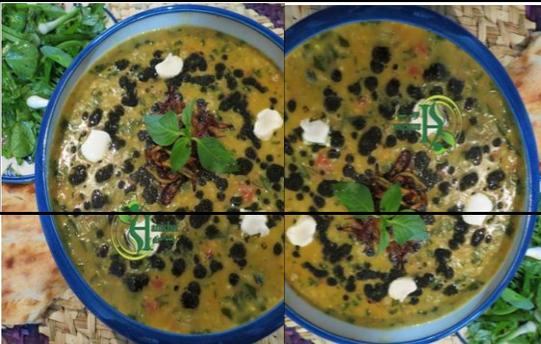|

Ingredients: Goat milk, chickpeas, rice, tomatoes, green beans, parsley, leek, basil, garlic, onion, salt, pepper and meat.

| | | Candies | | |
|---|---|---|---|---|
| Jelabi (جلبی یا جلوه ی) | Julavi | Jelabi (جلبی یا جلوه ی) | Джелаби | Jelabi (Yellow Sweet Circle Liquorices) |

**Explanation:** Jelabi is one of the delicious candies of Afghanistan that is mostly consumed during Ramadan. It is also a common gift for newly engaged couples during the Persian New Year (Nowruz).
Ingredients: Flour, water, oil, sugar, honey, cardamom, saffron, and yogurt.

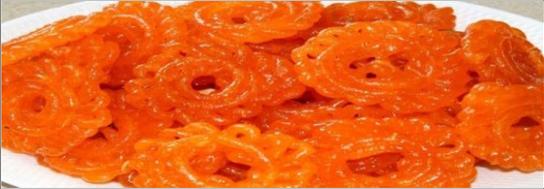

| | | | | |
|---|---|---|---|---|
| Konjetly Halva (کنجدلی حلوه) | Kunjutli Halva | Halva-e-Konjet (حلوای کنجد) | Халва с кунжутом (щербет с кунжутом) | Halva of Sesame |

**Explanation:** This candy is mostly used for breakfast. Most Afghan people prefer to eat it with tea. There is a district in the Faryab province of Afghanistan called Andkhoy that is popular for producing the most delicious Halva of Sesame in Afghanistan.

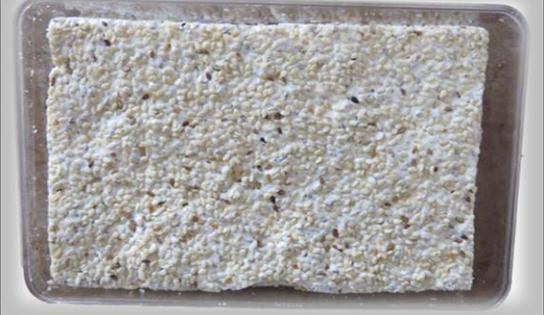

| | | Fruits | | |
|---|---|---|---|---|
| Tozli Nakhod (توزلی نخود) | Tuzli No'xat | Nakhod Doatasha Shor (نخود دوآتشه شور) | Солёный Горох {Солёный нод (горох)} | Salty Peas (Nakhod Doatasha Shor) |

**Explanation:** This is a dried fruit that is prepared with peas, it is mostly used on Eid days and it is very popular in Afghanistan, Iran, Pakistan, and India.

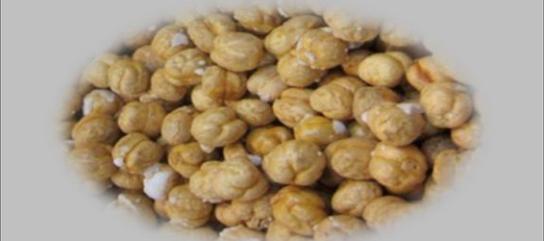

| | | Garments and Footwear | | |
|---|---|---|---|---|

| Guppi (گوپی) | Guppy / Ko'ylak | Gopicha (گیپچه) | Гупича (Мужская толстовка) | Gopicha |
|---|---|---|---|---|
| **Explanation:** This is traditional apparel worn in winter. Afghan tailors sew it. The clothes are used as a body shirt. It's very thick and many people during the winter wear it in order to keep warm. | | | | 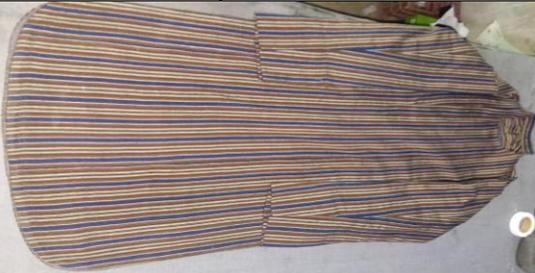 |
| Shelwar (شلور) | Shalvar | Shelwar (شلوار) | Щельвар (Брюки или щаравары) | Trousers (Outer Pants) or Shelwar |
| **Explanation:** These are traditional trousers or outer pants. This is worn with Gopi or separately. Both have the same cloth quality as Gopi which I explained at number 12. It is also used in winter when the weather gets very cold and is mostly used by villagers. | | | | 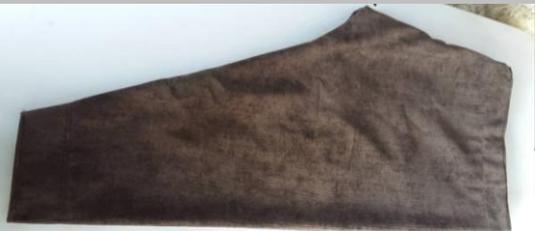 |
| Toon (تون) | Ton / Chopon | Chapan (چپن) | | |
| **Explanation:** This is an Afghan national and traditional garment, The Toon or Chapan is mostly worn by Uzbeks, Turkmen, Tajiks, and Pashtuns. This is like a long coat with arms and it is worn like outer cloths. | | | | 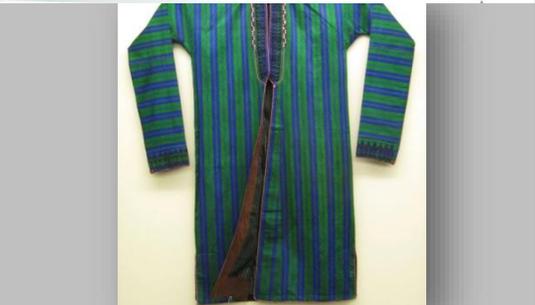 |

### Handicrafts, Industries etc…

| Khorjin (خورجین) | Xurjun | Khorjin (خورجین) | Хурджон (Седельная Сумка) | Saddlebag/bag (Khorjin) |
|---|---|---|---|---|
| **Explanation:** This is similar or close to a saddlebag which is put on a horse or donkey in order to carry things from different places. This is particularly used by those who live in villages. This is handmade. They put this kind of bag on a horse or donkey and then come to the city to buy materials. | | | | 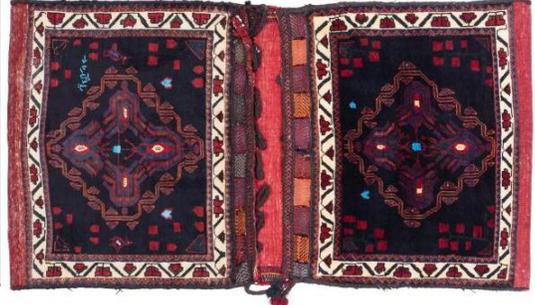 |

### Furnishings

| Boghcha (بوغچه) | Bug'cha | Boghcha (بوغچه) | Богча | Boghcha |
|---|---|---|---|---|

| **Explanation:** This is a piece of cloth where people pack their garments, this was used many years ago by most Afghan people, and at that time there were fewer cupboards but nowadays, it is rarely used by villagers. Its size is about 50/90cm. | 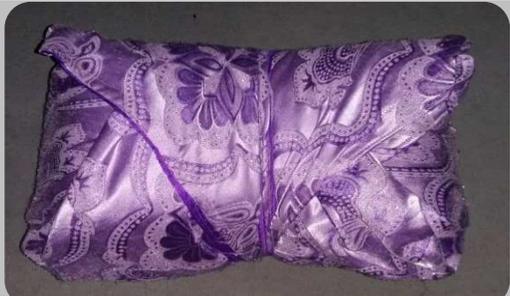 |
|---|---|

### Games

| Kurash (کوره ش) | Kurash | Pahlavani (پهلوانی) | Пахлавани (Борцы Кушти) | Pehlwani |
|---|---|---|---|---|

| **Explanation:** Pehlwani is a form of wrestling from the Indian subcontinent. It was developed in the Mughal Empire by combining native malla-buddha with influences from Persian varzesh-e bastani. The words pehlwani and kushti derive from the Persian terms pahlavani and kosht respectively. | 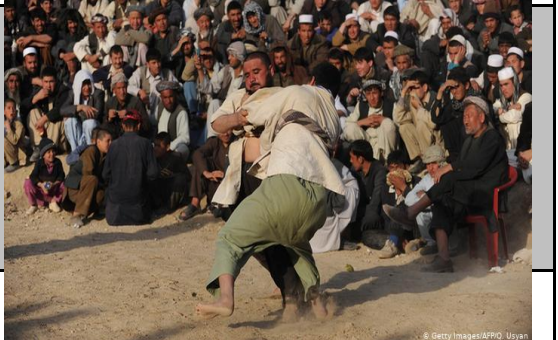 |
|---|---|

| Oghlaq (اوغلاق) | Uloq / Ko'pkari | Buzkashi (بزکشی) | | |
|---|---|---|---|---|

| **Explanation:** Buzkashi (literally "goat pulling" in Persian) is a Central Asian sport in which horse-mounted players attempt to place a goat or calf carcass in a goal. It is played mainly by communities in Central Asia (Uzbekistan, Tajikistan, Afghanistan, Kyrgyzstan, Turkmenistan, and Kazakhstan). Buzkashi is the national sport and a "passion" in Afghanistan where it is often played on Fridays and matches draw thousands of fans. Whitney Azoy notes in his book Buzkashi: Game and Power in Afghanistan that "leaders are men who can seize control by means foul and fair and then fight off their rivals. The Buzkashi rider does the same". Traditionally, games could last for several days, but in its more regulated tournament version, it has a limited match time. | 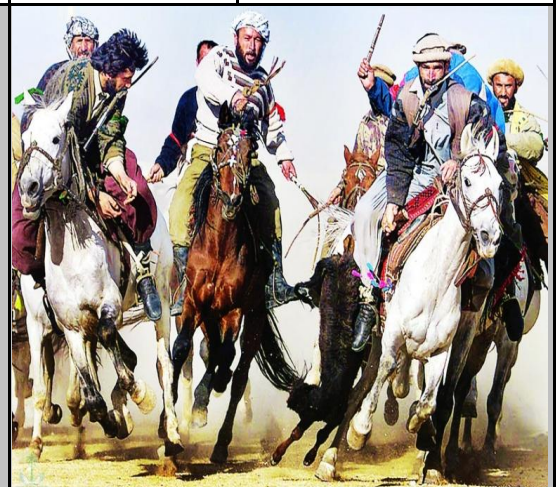 |
|---|---|
| **Explanation:** This game is played by children and adults, in this game each player digs a round hole (house) as the size of a small ball, and in the middle of houses they dig a round hole which is called a house of evil. Then one player throws the ball from a distance if the ball falls into the player's house (hole which is dug by the player) that player rides or sits on the shoulders of the player who is | 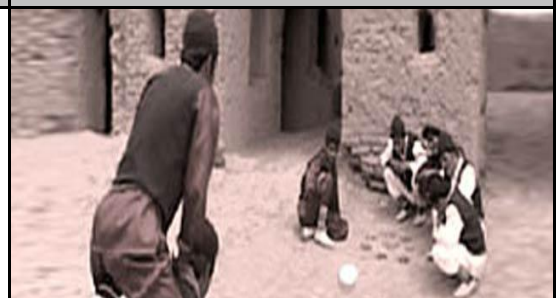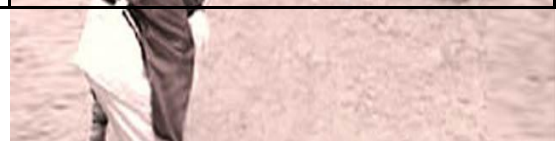 |

| | | | | |
|---|---|---|---|---|
| throwing the ball into houses. If the ball falls into the house of evil at that time other players run to the destination which is already determined and is about 30 meters far from the houses, the player who was throwing the ball into houses tries to strike and hit the players who are running to destination if the ball contacted to any player then that player continues the game and he throws the ball to houses. | | | | |
| Bish Tash (بیش تاش) | Besh Tash | Panj Saang (پنج سنگ) | Пандж Санг (Игра с Пятью Камешками) | Panj Saang (Game with Five Stones) |
| **Explanation:** This game is popular among Afghan females and it is one of the ancient games. This game needs five small stones and it can be played in couples or groups. The starter of the game is elected by lottery, the starter puts the stones on the rear of the hand and throws them up. When the stones prolapse the player must catch at least one of the stones. If she catches any of the stones, then she will repeat the process and throw those stones up and try to catch them again. The first person in the game to have no stones left will be out of the game (Qani, 2020, pp. 6. 28, 30, 32b) | | | 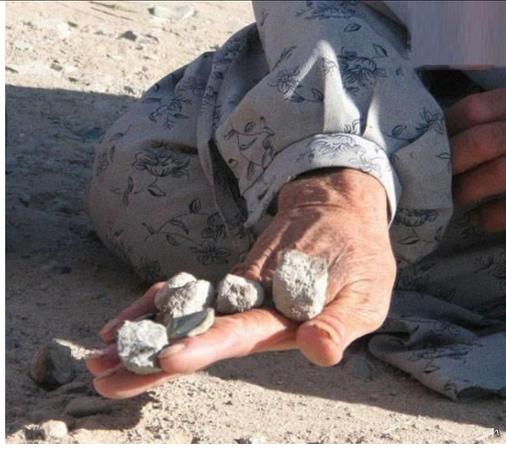 | |

## Conclusion

In conclusion, non-equivalent words refer to terms or phrases in different languages that do not have direct translations due to cultural, linguistic, or contextual differences. These words often carry unique meanings or convey emotions that cannot be fully captured in another language. Understanding the nuances of non-equivalent words can enhance cross-cultural communication and deepen our appreciation for the rich diversity of languages and cultures around the world.

This research paper includes more than 24 Uzbek and Dari words which had no direct equivalent in English and Russian Languages. It also indicates the theory and methods of rendering non-equivalent words into different languages. By reading this research paper the readers will learn the theoretical part of this research paper, Besides that, they will be familiar with interesting Dari and Uzbek words that have been professionally translated into different languages.

**چکیده**

ترجمه کلماتی که در زبان مقصد معادل ندارند کار آسانی نیست و یافتن معادل مناسب برای آن کلمات بسیار مهم است که قابل فهم بیان شود. زبان انگلیسی در سراسر جهان صحبت می‌شود و طبق آمار سال 2021، 1.35 میلیارد انگلیسی زبان و بیش از 258 میلیون روسی زبان وجود دارد. به ناچار، این میلیاردها سخنران در سراسر جهان با هم ارتباط دارند و ممکن است با موارد مختلفی سروکار داشته باشند. برای اینکه یکدیگر را بفهمند باید زبانی خالص و کاملاً قابل درک داشته باشند. این درک زبان خالص مستقیماً به دانش ترجمه مربوط می‌شود، جایی که زبان شناسان و مترجمان باید برای ریشه کن کردن سوء تفاهم کار و تحقیق کنند. سوء تفاهم ها بیشتر در کلمات غیرمعادل ظاهر می‌شود، زیرا در هر زبان واژه های محلی مختلفی مانند غذا، لباس، واژه های فرهنگی و سنتی و غیره وجود دارد. واقعاً اکثریت این کلمات در زبان مقصد معادل ندارند و بالای این کلمات باید کار صورت گیرد و معادل آن به زبان مقصد دریافت گردد، تا هر دو زبان به طور کامل قبل درک و فهم باشد. هدف از این تحقیق معرفی روش‌های حرفه‌ای تبدیل کلمات غیرمعادل از زبان مبدأ به زبان مقصد است و این تحقیق با استفاده از تحقیقات کتابخانه‌ای تکمیل شده است. با این حال، برخی از این کلمات غیرمعادل در حال حاضر به صورت حرفه ای به زبان مقصد ارائه شده اند، اما هنوز بسیاری از کلمات دیگر برای ارائه وجود دارد. در نتیجه، این مقاله تحقیقی شامل روش‌ها و قواعد متفاوتی برای ارائه واژه‌های غیرمعادل از زبان مبدأ به زبان مقصد است و 25 واژه غیرمعادل از زبان های اوزبیکی و دری به زبان‌های انگلیسی و روسی معادل سازی شده است.